*Chapter 7*

# Review of Data-Driven Generative AI Models for Knowledge Extraction from Scientific Literature in Healthcare

Leon Kopitar[1,2], Primoz Kocbek[1,3], Lucija Gosak[1] and Gregor Stiglic[1,2,4]

[1] *Faculty of Health Sciences, University of Maribor, 2000 Maribor, Slovenia.*
[2] *Faculty of Electrical Engineering and Computer Science, University of Maribor, 2000 Maribor, Slovenia*
[3] *Faculty of Medicine, University of Ljubljana, 1000 Ljubljana, Slovenia*
[4] *Usher Institute, University of Edinburgh, Edinburgh EH16 4SS, United Kingdom*
[*] *Author to whom correspondence should be addressed. E-mail: leon.kopitar1@um.si*

**Abstract.** This review examines the development of abstractive NLP-based text summarization approaches and compares them to existing techniques for extractive summarization. A brief history of text summarization from the 1950s to the introduction of pre-trained language models such as Bidirectional Encoder Representations from Transformer (BERT) and Generative Pre-training Transformers (GPT) are presented. In total, 60 studies were identified in PubMed and Web of Science, of which 29 were excluded and 24 were read and evaluated for eligibility, resulting in the use of seven studies for further analysis. This chapter also includes a section with examples including an example of a comparison between GPT-3 and state-of-the-art GPT-4 solutions in scientific text summarisation. Natural language processing has not yet reached its full potential in the generation of brief textual summaries. As there are acknowledged concerns that must be addressed, we can expect gradual introduction of such models in practise.

## 6.1 INTRODUCTION

Recent progress in Natural Language Processing (NLP) research using large pre-trained language models using Deep Neural Networks (DNN) where the number of parameters is in the order of 100 billion, commonly called Large Language Models (LLM), have pushed the limits of language understanding and text generation. Standard practice for such general-purpose models is to fine-tune them for task-specific downstream tasks, often via text prompts. This review presents a current overview of such models in healthcare with the downstream task of knowledge extraction from scientific literature, more specifically, abstractive summarization. Extracting very short and concise summaries of the paper content can save healthcare experts' time when browsing through the results of search queries, especially in cases when advice to the patient should be supported by the latest scientific literature immediately at the point of care. The review consists of a brief history of the development of the NLP summarization approaches and proceeds with familiarizing the reader with the methodology of the current State-of-the-art (SOTA) techniques. SOTA models are also examined, with a focus on models that can generate abstractive summaries, such as OpenAI's GPT-3 Curie and Davinci (Korngiebel and Mooney 2021; OpenAI 2022) with Too Long; Didn't Read (TLDR) summarization. These models are compared to existing models fine-tuned for extractive summaries in the healthcare domain, such as the Semantic Scholar Platform, which employs controlled abstraction for TLDRs with title scaffolding (CATTS) (Cachola et al. 2020). Opportunities and problems that arise with abstractive summaries are addressed. For example, one such problem is the evaluation of abstractive summaries since automatic methods, such as Recall-Oriented Understudy for Gisting Evaluation (ROUGE), that are extensively used for extractive summaries, for example, it was used for 2/3 of the summarization evaluations in summarization papers from North American Chapter of the Association for Computational Linguistics (NAACL) and Association for Computational Linguistics (ACL) in 2021 (Kasai et al. 2022), but might not be a viable option when using LLM. Current trends and best practice guidelines are still in favour of human evaluations (van der Lee et al. 2021). However, some suggestions of possible reference-free metrics for NLG evaluations, such as G-Eval (Liu et al. 2023) and GPTScore (Fu et al. 2023), are applicable to new tasks without human references and might eventually replace human evaluation.

The review covers the following topics: Introduction to a brief history of text summarisation using NLP, SOTA NLP summarization approaches, applications in the healthcare domain, showcased evaluations of abstractive summaries generated by SOTA models, limitations, and current challenges of existing methods. We see an additional value of such a review in providing important feedback for developers of such models for future improvements.

### 6.1.1 Introduction to a brief history of text summarisation using NLP

The history of text summarization begins with the development of NLP. The origin of NLP is usually attributed to Machine Translation, which was used during the second world war to translate the English language to Russian and the other way around (Sreelekha et al. 2016).

During the 1950s and 1960s, Luhn and Edmundson have been studying ways of automatically extracting documents using different approaches: the topic is hidden within initial sentences (e.g. in the first paragraph, immediately after sections such as "Introduction", "Purpose", "Conclusions"), topic sentences are located with the presence of pragmatic words ("significant", "impossible", "hardly") and location-wise, where the key information is hidden in the first and the last sentence of paragraphs (Luhn 1958; Edmundson 1969).

The rise of text summarization appeared in the 1980s when researchers perceived NLP as an opportunity for a topic for research. These days, a rule-based algorithm known as an *importance evaluator* was predominantly used in text summarization. The importance evaluator used two knowledge bases, the importance rule base that contained knowledge conveyed through IF-THEN rules, and "encyclopedia" containing domain-specific world knowledge constituted of a network of frames (Fum et al. 1986).

Later, the development of text summarization progressed with the introduction of the Diversity-based approach in extractive summarization. This approach calculated the diversity of sentences and removed redundant sentences from the final abstract. Diversity was calculated using the K-means clustering algorithm with Minimum Description Length Principle (Nomoto and Matsumoto 2001).

Another way to extract sentences was to apply Graph-based ranking algorithms. These algorithms determine the importance of a node within a graph according to global information computed recursively from the entire graph. TextRank sentence extraction algorithm uses a similar principle, instead, the graph is built from natural language text and includes multiple or partial links, represented as the graph's edges, that are extracted from the text (Mihalcea 2004).

In 2009, Suanmali et al. (2009) introduced text summarization based on the general statistic method (GSM) and fuzzy logic method to decide on the importance of the text. This method consisted of four components: fuzzifier, inference engine, defuzzifier, and fuzzy knowledge base.

A few years later, the encoding and decoding model was developed: Sequence-to-sequence (Seq2Seq) learning model. The architecture of Seq2Seq was composed of an encoder, a Recurrent neural network (RNN) with LSTM/GRU, and a decoder (RNN). The role of an encoder was to encode the input sequence into a single fixed-size vector, while the decoder yielded an output sequence based on the fixed-size vector. Performance issues with Seq2Seq arose when long and complex sentences were tried to be encoded into a single fixed-size vector (Sutskever et al. 2014). This was later resolved with an Attention Mechanism technique, which identified relevant input tokens, and elements of the text, by computing context vectors for all tokens in the input sentence for each token in the output sentence (Bahdanau et al. 2015). Note that tokens are basic unit in many LLM, usually defined as groups of characters and used to compute the length of text, for example in GPT models 1000 tokens represent approximately 750 words and a paragraph usually consists of around 35 tokens (OpenAI 2022).

Based on the achieved improvements with the Attention Mechanism technique, Vaswani et al. (2017) proposed Transformer architecture, which uses self-attention that selectively focuses on parts of the input, and next, multi-head attention that deals with multiple parts simultaneously. Furthermore, it does not use recurrence or convolution and it relies on position-wise feed-forward network architecture (Vaswani et al. 2017).

The evolution of pre-trained language models began with the introduction of Bidirectional Encoder Representations from Transformer (BERT). BERT is based on transformer architecture, using Masked Language Modelling (MLM) and it is pre-trained on a vast amount of text data, such as articles, books, and other literature. Reportedly it is pre-trained on over 3.3 billion words (Devlin et al. 2019). Like other LLM that will be introduced later, BERT uses a transfer learning technique, specifically sequential transfer learning that consists of two steps: the pre-training stage, where NN is trained on general data, and fine-tuning stage where the model is trained on domain-specific task (Zhang et al. 2019; Devlin et al. 2019).

In the same year, the corporation OpenAI developed the first one in a series of Generative Pre-training Transformers (GPT), known as GPT version 1 (GPT-1) (Radford et al. 2018). A year later GPT-2 (with 1.5 billion parameters) (Alec et al. 2019), and GPT-3, in 2020 (Brown et al. 2020). GPT-3 is pre-trained on 100 times more parameters than its predecessor, precisely on 175 billion parameters (Brown et al. 2020).

During the development of the LLM series by OpenAI, in 2020, Google's teams introduced the PEGASUS model that utilizes MLM together with gap sentence generation (GSG). Their model removes/masks important sentences from the input text and generates new sentences from the remaining sentences. PEGASUS generates great results with as many as 1000 examples (Zhang et al. 2020b).

In late 2022, ChatGPT, another variant of GPT was released. ChatGPT is used as a chatbot built on top of GPT3.5 and has the ability to provide short abstracts on demand (Aydın and Karaarslan 2022; Jiao et al. 2023). It should be noted, that in the first quarter of 2023 GPT-4 was released, and although no specific details of the model are known (OpenAI 2023), but early indications do imply improvements in multiple downstream tasks compared to previous iterations.

## 6.2 METHODS

In this section, we introduce the basic terminology relevant to short summarization models and study selection for this review.

### 6.2.1 Extractive and abstractive summarization

Extractive summarization is a technique that extracts the most important sentences from the text and arranges them into a logical summary (Gupta and Gupta 2019) without generating any new sentences nor altering the original text. Abstractive summarization is a technique that summarises text by understanding the content and produces new or rephrased sentences of the original text. It is considered to be one of the more challenging tasks in NLP since it combines the understanding of long passages, information compression, and language generation (Zhang et al. 2020a). The aim of the abstractive summarization is to produce human-like summaries that successfully capture the meaning of the source text.

Challenges will be addressed in the subsection "2.3.4 Limitations and challenges of existing SOTA models".

### 6.2.2 Zero-shot

Text summarization is performed mainly with two types of meta-learning: zero-shot learning and few-shot learning. Zero-shot learning is associated with the process when learners can resolve tasks that have not been presented to the learner before (Kojima et al. 2022) and is solely based on the description of the tasks (Romera-Paredes and Torr 2015).
One study indicated that LLM, specifically GPT-3 performed well at zero-shot information extraction from clinical text despite not being trained specifically for the clinical domain (Agrawal et al. 2022).

### 6.2.3 Few-shot learning

On the other hand, few-shot learning (Bataa and Wu 2020; Chintagunta et al. 2021), supplies the learner with unseen samples of data to learn a new task (Kojima et al. 2022). In scenarios when few-shot learning is utilized, it is often sufficient to apply only a couple of examples. Speaking of meta-learners, Brown et al. demonstrated that zero-, one-, and few-shot performance increases with the increase of model capacity, indicating that larger models are more accomplished (Brown et al. 2020).
In one of the previous studies, researchers showed that the few-shot GPT-3 model performs poorly in the biomedical domain, specifically, in experiments that deal with textual inference, estimating semantic similarity, question answering, and others (Moradi et al. 2021).

### 6.2.4 Search strategy

Our general research question was to investigate to what extent LLM can be used to extract viable knowledge in healthcare settings. The following search engines were used for conducting a scoping review: PubMed, Web of Science, CINAHL/Medline. We limited the review to the studies that were conducted in the past five years (2017 - 2022). Studies were retrieved in December 2022 using the following search term:

((((text OR abstract OR scientific literature OR scientific document) AND (summarization OR summarisation OR TLDR)) AND (large language models OR LLM OR NLP OR natural language processing) AND (healthcare OR health care OR primary care OR patient care OR nursing OR medical care)).

### 6.2.5 Study selection

The screening was carried out according to the guidelines of Arksey and O'Malley (2005). In the first step, duplicate studies were removed. The remaining studies were assessed by checking the title and abstract for the presence of words such as "summari(s/z)ation", "TLDR", terms that relate to healthcare, and phrases that resemble and belong to the natural language processing domain. Systematic and other reviews were included and reviewed regardless of the satisfaction of the above criteria. Included studies were then skimmed and reviewed for relevance. We have excluded research in languages other than English. Studies that met those given criteria underwent full-text reading individually by two authors. In case of disagreement, the authors resolved their differences through discussion.

The study selection process is also displayed in the result section using a PRISMA flow (Page et al. 2022). Additionally, In the Results section, we demonstrate some examples of evaluations of abstract summaries generated by SOTA models.

## 6.3 RESULTS

### 6.3.1 Search results

Following PRISMA diagram (Figure 2.1), a total of 60 studies were identified in PubMed (45) and Web of Science (15). After the removal of duplicated records, 54 studies underwent a screening process. During the initial step of screening, 29 studies were excluded since either the title or the content of the abstract did not cover desired topics described in the subsection "Study selection", and one study could not be retrieved. In the next step, 24 studies were fully read and assessed for eligibility, where seven reviews did not cover relevant topics, and eight studies reported only keyword/symptom identification or extraction without resulting in any form of text summaries. Additionally, two studies were excluded due to summarising connections in a form of tabular summarization or their identification. Thus, the selection process yielded seven studies that were relevant to the review process.

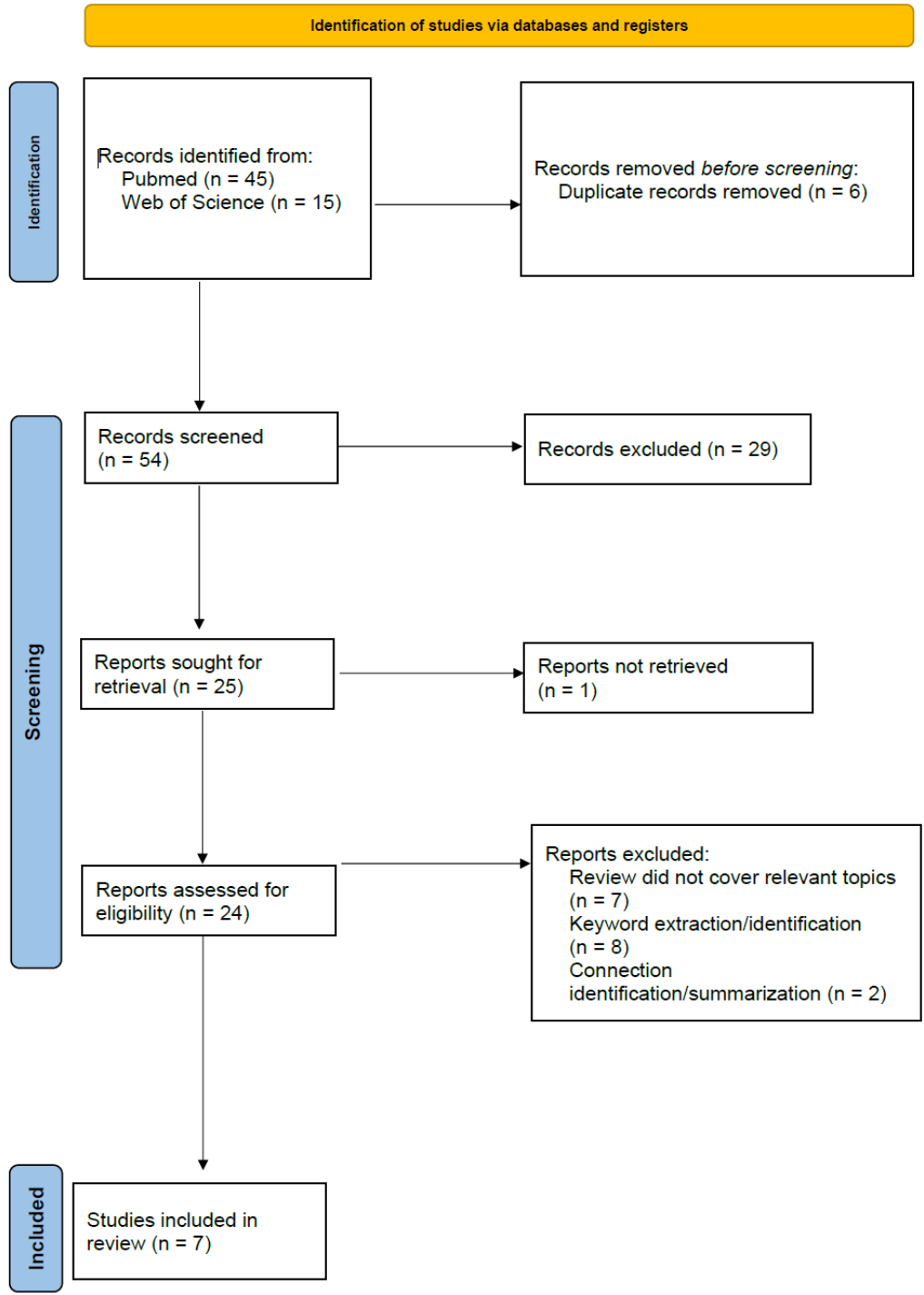

**Figure 2.1**. PRISMA diagram (Page et al., 2021)

Data-Driven Knowledge Extraction from Healthcare Literature        7*2.3.1.1 Characteristics of the included studies*

All included studies were published as a scientific article either as a part of conference proceedings (4) or journals (3). One study was published in 2017, 2018 and in 2020, while five studies were published in 2022. Almost half of the included studies were conducted in the USA (42%), remaining studies were based in Europe (Table 2.1.). Four studies included extractive text summarization in their research, while three studies dealt with abstractive summarization.

**Table 2.1.** Characteristics of the included studies

| Authors | Year | Country | Category | Data source | Type of text summarization /generation | Type of learning | Goal |
|---|---|---|---|---|---|---|---|
| **Sadeh-Sharvit et al.** | 2022 | USA | Retrospective study | Clinical session | Extractive | N/A | Examine the use of session summaries |
| **Trivedi et al.** | 2018 | USA | Research article | Signout notes | Extractive | Semi-supervised | New tool proposal for Signout note preparation |
| **Lee and Uppal** | 2019 | USA | Research article | Clinical / biomedical text | Extractive | Mutli stage process: TF-IDF, Random forests and indicators of importance | Proposing CERC system built |
| **Goldstein et al.** | 2017 | Israel | Research article | Clinical records | Abstractive | Few-shot | Examine automated creation of meaningful summaries |
| **Kocbek et al.** | 2022 | Slovenia | Research article | Healthcare literature | Abstractive | Zero-shot and few-shot | Explore the use and usefulness of several SOTA models for generating short summaries |
| **Stiglic et al.** | 2022 | Slovenia | Case study | Healthcare literature | Abstractive | Zero-shot and few-shot | Use case scenario of SOTA models for generating short summaries in healthcare |
| **Reunamo et al.** | 2022 | Finland | Research article | Nursing entries | Extractive | LSTM-based extraction using LIME explainability | Explore keyword-based text summarization method based on |

|  | ML model explainability |
|---|---|

### 6.3.2 Applications in the healthcare domain

In the study conducted by Sadeh-Sharvit et al. (2022), an AI therapy-specific platform (Eleos health platform) was proposed. This platform uses speech-to-text technology by translating what is being said into suggestions for progress note documentation that allows for generating structured sentence-by-sentence summaries (Sadeh-Sharvit et al. 2022).

Lee and Uppal developed a multi-indicator text summarization algorithm (MINTS) for generating extractive summaries from clinical and biomedical texts. MINTS uses random forests classifiers and other indications like sentence length, the position of input text, number and percentage of clinical/biomedical terms, normalized degree centrality, and overlap with global term frequency distribution calculated by using a similarity metric known as Sørensen–Dice-coefficient/index (DS) (Lee and Uppal 2020). Compared to other extractive summarization ranking methods (topicDist (Zhang et al. 2020a), LexRank (Erkan and Radev 2011), Position-based ranking, and Random selection), MINTS achieved the highest evaluation score. Additionally, they proposed a web-based interactive summarization and visualization tool (CERC), which uses indicative, as well as informative summarization techniques (Lee and Uppal 2020).

Goldstein et al. proposed an automatic summarisation system (CliniText - CTXT) based on Intensive Care Unit (ICU) clinical knowledge base that generates summaries that can serve as a base for a discharge letter. The system was evaluated by relative completeness, several quality parameters (such as readability, comprehensiveness, supportive …), functional performance, and correctness of clinicians' answers (Goldstein et al. 2017). CTXT generates summaries of longitudinal clinical data by combining knowledge-based temporal data abstraction, textual summarization, abduction, and NLP (Goldstein and Shahar 2013).

In another study, researchers presented an interactive approach to help clinicians to identify relevant text for inclusion in sign-out notes. Their tool, NLPReViz, uses extractive summarization for providing suggestions that can then be accepted or rejected by the user (Trivedi et al. 2018).

In 2022, Kocbek et al. explored the use and usefulness of several SOTA models, such as OpenAI Davinci, OpenAI Curie, Pegasus-XSum, and BART-SAMSum for generating short summaries from abstracts of healthcare scientific literatures (Kocbek et al. 2022). Later, the usefulness and relevance of generated abstractive short summaries were also evaluated in the use case scenario (Stiglic et al. 2022). Abstractive summaries from the Semantic Scholar platform (Hannousse 2021), which served as a baseline model, achieved the highest average score in the evaluation metrics (Naturalness, Quality, and Informativeness) (Kocbek et al. 2022).

In yet another research, authors introduced a keyword-based text summarization method for nursing entries in EHRs using model explainability (Reunamo et al. 2022).

### 6.3.3 Evaluation of generated short summaries

In the selected studies, researchers carried out an evaluation of generated summaries using various techniques. Lee et al. evaluated generated texts using ROUGE metrics, in particular ROUGE-1, ROUGE-2, and ROUGE-SU4, comparing generated text with the reference text according to overlapping unigrams, bigrams, and n-grams with the maximum skip distance of 4, respectively. The ROUGE metric is one of the most widely evaluation techniques used in general (Lin and Och 2004; Yan et al. 2011).

The study of Reunamo et al. (2022) reported using a manual evaluation scale of four classes, that evaluates the quality of information being conveyed. Similarly, Kocbek et al. (2022) used "Informativeness" as one of the evaluation measures.

Measures like "Readability", "Comprehensiveness", "Relevance", "Naturalness" and "Quality", were applied only once through all seven studies, where "Readability" and "Comprehensiveness" were included in the study of Goldstein et al., while among others, "Naturalness" and "Quality" were present in the study of van der Lee et al. (2021) and Kocbek et al. (2022).

The assessment criteria can also encompass other quality measures, such as clinical course and continuity of care. The clinical course measures the effectiveness of the text in enabling clinicians to understand the events and experiences of the patient throughout their ICU stay, meanwhile the continuity of care measures the extent to which the information provided in the text supports the flowless continuation of care (Goldstein et al. 2017).

### 6.3.4 Examples of evaluations of summaries generated by SOTA models

The first example is from Reunamo et al. (2022) and is called an Explainer Extractor, which extracts keywords and keyphrases through explainable AI, more specifically it combines a text classification model (bidirectional LSTM) with model explainability (local interpretable model-agnostic explanations (LIME)). An example can be seen in Figure 2.2, where thou it might not provide a full sentence, it does provide a reasonable explanation and could we have turned into sentences with advanced transformer-based methods. First part shows the keywords that were recognized as coefficients of LimeTextExplainer with the highest absolute values. Coefficients were then Z-standardized and weighted by paragraph score and coloured accordingly. In the second part, keywords that appeared consecutively were merged together to form keyphrases, and the keyphrase scores were determined based on the highest scores of the individual components. The last step shows the result after the removal of stop words and duplicate keywords.

| *The significance of keywords correlates with the shade of the token's background color, transitioning gradually from lighter to darker shades of green* |
|---|
| *Regular exercise is widely known to have numerous benefits for overall health. It helps improve cardiovascular health, strengthens muscles and bones, boosts mood, and can even aid in weight management. Incorporating physical activity into daily routines, such as walking, cycling, or swimming, can significantly contribute to a healthier lifestyle and reduce the risk of chronic conditions like heart disease, diabetes, and obesity. Moreover, exercise promotes better sleep and increases energy levels, making it a cornerstone of preventive healthcare.* |
| *Adjacent keywords are merged into keyphrases when they occur together.* |
| *Regular exercise is widely known to have numerous benefits for overall health. It helps improve cardiovascular health, strengthens muscles and bones, boosts mood, and can even aid in weight management. Incorporating physical activity into daily routines, such as walking, cycling, or swimming, can significantly contribute to a healthier lifestyle and reduce the risk of chronic conditions like heart disease, diabetes, and obesity. Moreover, exercise promotes better sleep and increases energy levels, making it a cornerstone of preventive healthcare.* |
| *Keyword summary* |
| *Regular exercise; helps improve cardiovascular health; strengthens muscles and bones, boosts mood; aid in weight management; healthier lifestyle; reduce the risk of chronic conditions; better sleep and increases energy levels* |

Figure 2.2. Word scoring example of a paragraph

Next, we look at a qualitative example from the abstract by Stine et al. (2021) and compare the results of four different LLMs (Figure 2.3). Semantic scholar summarized by extracting information directly from study conclusion. OpenAI Davinci summarized an abstract into a short statement without any supportive information. Interestingly, OpenAI Curie included a sentence that includes the guidelines of The American College of Sports Medicine (ACSM), which is not even mentioned in the original abstract, while completely ignoring non-alcoholic fatty liver disease (NAFLD). Alike Semantic Scholar, PEGASUS-Xsum generated short summary by extracting the content from the results.

**Figure 2.3.** Qualitative evaluation of short summaries with annotations (1) (Stiglic et al., 2022)

Another example of the qualitative evaluation is based on the abstract by Kanellopoulou et al. (2021) (Figure 2.4). Here, Semantic scholar (SS_tldr_baseline) generated short summary mainly from the objective and conclusions. While in the previous example OpenAI Davinci provided a general statement, this time it extracted the beginning of the result. Similarly, to the previous example, OpenAI Curie (OpenAI_tldr_Curie) added an additional information to this generated short summary as well. This time it clearly expressed its opinion by the addition of the following sentence: "I'm not sure what to make of this study". Pegasus-Xsum summarized the abstract in a very general manner, while Bart-SAMSum extracted a part of the results without any supporting numbers.

**Figure 2.4.** Qualitative evaluation of short summaries with annotations (2) (Kocbek et al., 2022)

Below we provide two examples of improvements in short summarizations using GPT-4 compared to GPT-3 and older models (Figure 2.5), in the first case the improvement seems minor (Figure 2.5a), whereas in the second case there is a major improvement (Figure 2.5b).

In the example of the study by Spaulding et al. (2021) (Figure 2.5a), Semantic scholar and GPT-4 were relying on extractive summarisation technique. While Semantic scholar provided summary directly from the conclusions, GPT-4 extended that part by unabbreviating a term and pointing out the type of study that was conducted. One of the LLMs, OpenAI Davinci included a sentence that resembles a reviewer comment, while Pegasus-Xsum supplied a false information. Another example (Figure 2.5b), which is shown on the study by Gutierrez et al. (2021), OpenAI Davinci, Curie and Pegasus-Xsum achieved much lower average relevance score than Semantic scholar. Enriched and improved summarisation in comparison to GPT-3 models.

**a.**

**Abstract 10 (Flesch reading ease score 31.96*)**

*Background* The use of mobile health (mHealth) interventions, including smartphone apps, for the prevention of cardiovascular disease (CVD) has demonstrated mixed results for obesity, hypercholesterolemia, diabetes, and hypertension management. A major factor attributing to the variation in mHealth study results may be mHealth user engagement. *Objective* This systematic review aims to determine if user engagement with smartphone apps for the prevention and management of CVD is associated with improved CVD health behavior change and risk factor outcomes. *Methods* We conducted a comprehensive search of PubMed, CINAHL, and Embase databases from 2007 to 2020. /.../ The quality of the studies was assessed. *Results* Of the 24 included studies, 17 used a randomized controlled trial design, 4 used a retrospective analysis, and 3 used a single-arm pre- and posttest design. //.../Most of the studies that assessed the relationship between user engagement and reduction in weight (9/13, 69%), BMI (3/4, 75%), body fat percentage (1/2, 50%), waist circumference (2/3, 67%), and hemoglobin A1c (3/5, 60%) found statistically significant results, indicating that greater app user engagement was associated with better outcomes. /.../ *Conclusions* Increased app user engagement for prevention and management of CVD may be associated with improved weight and BMI; however, only a few studies assessed other outcomes, limiting the evidence beyond this. Additional studies are needed to assess user engagement with smartphone apps targeting other important CVD risk factors, including dietary behaviors, hypercholesterolemia, diabetes, and hypertension./.../.

**Semantic Scholar TLDR** Relevance: 4.44 (SD=1.59)

Increased app user engagement for prevention and management of CVD may be associated with improved weight and BMI; however, only a few studies assessed other outcomes, limiting the evidence beyond this.

**OpenAI Davinci** Relevance: 5.11 (SD=1.62)

The evidence is mixed, but there is a trend toward greater engagement with smartphone apps for the prevention and management of CVD being associated with better outcomes. This is a systematic review, so the quality of the studies is important. The authors have done a good job of identifying the limitations of the studies, and they have done a good job of summarizing their findings.

**OpenAI Curie** Relevance: 5.78 (SD=1.09)

There is a lot of variation in the quality of the studies, but the overall picture is that there is a positive relationship between mHealth user engagement and CVD risk factor outcomes.

**PEGASUS Xsum** Relevance: 3.78 (SD=1.56)

Increased user engagement with a smartphone app may be associated with improved weight loss and BMI, according to a systematic review and meta-analysis of 24 studies.

**GPT-4**

This systematic review found that increased user engagement with smartphone apps for the prevention and management of cardiovascular disease (CVD) may be associated with improved weight and BMI. However, only a few studies assessed other outcomes, limiting the evidence beyond this. Additional research is needed to assess user engagement with smartphone apps targeting other important CVD risk factors, including dietary behaviors, hypercholesterolemia, diabetes, and hypertension.

*FRE: 30-50 difficult to read, best understood by college graduates*

**b.**

**Abstract 20 (Flesch reading ease score 40.77*)**

Multi-morbidity greatly increases the complexity of managing disease, particularly among underserved populations who also suffer from mental health disorders. Multi-morbidity is also more likely to worsen outcomes and reduce positive behavioral changes due to coexisting depression or anxiety, complicating global/individual health care needs. The objective of this study was to determine the prevalence of having three or more chronic conditions in Latinx patients receiving medical care at Caridad Center who suffer from depression and anxiety.
 A needs assessment was conducted on Latinx patients who sought care for mental health from Caridad Center in Boynton Beach, FL to determine which co-morbidities they experience. Medical records of 100 random patients diagnosed with depression and/or anxiety were /..../
 The mean age was 51.9 ± 11.9 years and majority were female, 79.4%. Out of the 100 individuals, 43% were diagnosed with depression, 38% with anxiety and 17% with both depression and anxiety. In those with depression, 54% had ≥3CVD risk factors, 39% had ≥3 diagnosed co-morbidities, 47% had hypertension and hyperlipidemia, and 38% had type 2 diabetes. Additionally, 43% of those with anxiety had ≥3 CVD risk factors, 31% ≥3 diagnosed co-morbidities, 46% had hypertension, 41% had hyperlipidemia and 31% had type 2 diabetes. /..../
 Completion of a needs assessment from Caridad Center, the largest provider of free medical and mental health services in Florida, demonstrated that multi-morbidity were prevalent among the Latinx sample afflicted by depression and/or anxiety. More work is needed to further investigate how lifestyle and other risk factors interplay with the severity of mental illness in this population.
FIU RCMI/NIMHD.

**Semantic Scholar TLDR** Relevance: 4.67 (SD=1.80)

Completion of a needs assessment from Caridad Center demonstrated that multi-morbidity were prevalent among the Latinx sample afflicted by depression and/or anxiety, and more work is needed to further investigate how lifestyle and other risk factors interplay with the severity of mental illness in this population.

**OpenAI Davinci** Relevance: 3.67 (SD=1.87)

Latinx with depression and/or anxiety have a high prevalence of multi-morbidity.

**OpenAI Curie** Relevance: 3.22 (SD=2.11)

Latinx people are more likely to suffer from depression and anxiety.

**PEGASUS Xsum** Relevance: 3.22 (SD=1.79)

  Multi-morbidity is one of the most common co-morbidities among patients with mental health disorders.

**GPT-4**

A study conducted on Latinx patients at Caridad Center in Florida found that multi-morbidity is prevalent among those suffering from depression and/or anxiety. The majority of the patients were female, and many had multiple chronic conditions, such as hypertension, hyperlipidemia, and type 2 diabetes. The study highlights the need for further investigation into how lifestyle and other risk factors affect the severity of mental illness in this population.

*FRE: 30-50 difficult to read, best understood by college graduates*

**Figure 2.5.** Examples of qualitive improvements in short summarization using GPT-4 (March 2023 version) with annotations.

### 6.3.5 Limitations and challenges of existing SOTA models

Models generate repetitive and generic summaries especially when input text is very long. Summarization of very long documents present a problem of delivering coherent summaries (Gupta and Siddiqui 2012). At times, models can be misled by metaphors or idiomatic expressions and change the meaning of text during summarization process. Another issue is that these models are pre-trained on English text data and might not perform well in text summarization in other than non-global languages (Aksenov et al. 2020).

Content from images can be extracted using various methods, such as Optical Character Recognition (OCR), which extracts text from images (Indravadanbhai Patel et al. 2012), or image captioning models, which can generate descriptions of images. Features such as relationships or patterns are usually not interpretable and understandable by the model, which can affect the quality of the generated summary (Fan et al. 2018).

One of the main limitations is the scalability of human evaluations, both in terms of time and costs, but are still favoured in current trends and best practice guidelines (van der Lee et al. 2021).

## 6.4 DISCUSSION

This review explores the findings and implications of the research conducted on the application of SOTA models in text summarization within the healthcare domain. It delves into the various approaches and techniques adopted by researchers.

This study observed that only a few studies utilized models, such as those that are based on GPT-3 for the purpose of text summarisation in a healthcare domain. Moreover, most included studies dealt with an extractive summarization. Consequently, there exists a lack of in-depth exploration of abstractive summarisation, that provides paraphrased, concise, and coherent summaries. Studies included in this review focus on generating summaries from clinical, biomedical texts, speech-to-text generated text, and nursing entries in EHRs. Based on that, there is a potential in research dealing with radiology reports, pathology reports, and similar. Another area, that can be focused on, is in establishing standardized evaluation metrics for assessing the effectiveness of methods for generating short summaries. Since some studies mentioned metrics such as readability, completeness, other naturalness, quality, and informativeness. Summary evaluation in healthcare lags behind novel evaluation approaches that are proposed for LLM. In general, studies evaluate summaries with limited, single aspects (fluency, readability …) or multi-aspects but neglect their definition or even more relationship among aspects. Even more, such evaluations require time-consuming manual annotations of samples. These issues were divulged by Fu et al. (2023), who proposed a novel network GPTScore that utilizes NLP instructions (zero-shot instructions) and contextual learning to overcome multiple evaluation challenges simultaneously (Fu et al. 2023). Another approach to summary evaluation was introduced by Liu et al. They proposed G-Eval that uses LLMs with chain-of-thoughts (CoT) (Wei et al. 2022; Liu et al. 2023). It requires a prompt with the definition of the evaluation task and criteria which is a collection of intermediary instructions (CoT) generated by LLM that delineate the precise sequence of evaluation steps (Liu et al. 2023). The difference in evaluation scoring between these two is that GPTScore uses probability in text generation as an evaluation metric, whereas G-Eval performs the evaluation with a form-filling paradigm.

The use of NLP for short text generation has not reached its peak potential. Due to the known issues, it is still unclear when the need for human evaluation could be replaced by an LLM.

## 6.5 REFERENCES

Agrawal M, Hegselmann S, Lang H, et al (2022) Large language models are few-shot clinical information extractors. 1998–2022

Aksenov D, Moreno-Schneider J, Bourgonje P, et al (2020) Abstractive Text Summarization based on Language Model Conditioning and Locality Modeling. In: Proceedings of the Twelfth Language Resources and Evaluation Conference. European Language Resources Association, Marseille, France, pp 6680–6689

Alec R, Jeffrey W, Rewon C, et al (2019) Language Models are Unsupervised Multitask Learners | Enhanced Reader. OpenAI Blog 1:9

Arksey H, O'Malley L (2005) Scoping studies: towards a methodological framework. https://doi.org/101080/1364557032000119616 8:19–32. https://doi.org/10.1080/1364557032000119616

Aydın Ö, Karaarslan E (2022) OpenAI ChatGPT Generated Literature Review: Digital Twin in Healthcare. SSRN Electron J. https://doi.org/10.2139/ssrn.4308687

Bahdanau D, Cho KH, Bengio Y (2015) Neural machine translation by jointly learning to align and translate. 3rd Int Conf Learn Represent ICLR 2015 - Conf Track Proc

Bataa E, Wu J (2020) An investigation of transfer learning-based sentiment analysis in Japanese. ACL 2019 - 57th Annu Meet Assoc Comput Linguist Proc Conf 4652–4657. https://doi.org/10.18653/v1/p19-1458

Brown TB, Mann B, Ryder N, et al (2020) Language models are few-shot learners. Adv Neural Inf Process Syst 2020-Decem:

Cachola I, Lo K, Cohan A, Weld DS (2020) TLDR: Extreme Summarization of Scientific Documents. Find Assoc Comput Linguist Find ACL EMNLP 2020 4766–4777. https://doi.org/10.18653/V1/2020.FINDINGS-EMNLP.428

Chintagunta B, Katariya N, Amatriain X, Kannan A (2021) Medically Aware GPT-3 as a Data Generator for Medical Dialogue Summarization. Proc Mach Learn Res 126:1–18

Devlin J, Chang MW, Lee K, Toutanova K (2019) BERT: Pre-training of deep bidirectional transformers for language understanding. NAACL HLT 2019 - 2019 Conf North Am Chapter Assoc Comput Linguist Hum Lang Technol - Proc Conf 1:4171–4186. https://doi.org/10.48550/arxiv.1810.04805

Edmundson HP (1969) New Methods in Automatic Extracting. J ACM 16:264–285. https://doi.org/10.1145/321510.321519

Erkan G, Radev DR (2011) LexRank: Graph-based Lexical Centrality as Salience in Text Summarization. J Artif Intell Res 22:457–479. https://doi.org/10.1613/jair.1523

Fan C, Zhang Z, Crandall DJ (2018) Deepdiary: Lifelogging image captioning and summarization. J Vis Commun Image Represent 55:40–55. https://doi.org/10.1016/J.JVCIR.2018.05.008

Fu J, Ng S-K, Jiang Z, Liu P (2023) GPTScore: Evaluate as You Desire. arXiv Prepr arXiv230204166

Fum D, Guida G, Lasso(') C" (1986) Tailoring Importance Evaluation to Reader's Goals: A Contribution to Descriptive Text Summarization

Goldstein A, Shahar Y (2013) Implementation of a system for intelligent summarization of longitudinal clinical records. Lect Notes Comput Sci (including Subser Lect Notes Artif Intell Lect Notes Bioinformatics) 8268 LNAI:68–82. https://doi.org/10.1007/978-3-319-03916-9_6/COVER

Goldstein A, Shahar Y, Orenbuch E, Cohen MJ (2017) Evaluation of an automated knowledge-based textual summarization system for longitudinal clinical data, in the intensive care domain. Artif Intell Med 82:20–33. https://doi.org/10.1016/j.artmed.2017.09.001

Gupta S, Gupta SK (2019) Abstractive summarization: An overview of the state of the art. Expert Syst Appl 121:49–65. https://doi.org/10.1016/J.ESWA.2018.12.011

Gupta VK, Siddiqui TJ (2012) Multi-document summarization using sentence clustering. 4th Int Conf Intell Hum Comput Interact Adv Technol Humanit IHCI 2012. https://doi.org/10.1109/IHCI.2012.6481826

Gutierrez M, Diaz-Martinez J, Maisonet J, et al (2021) Co-morbid Health Conditions in Latinx Adults Receiving Care for Depression and Anxiety. Curr Dev Nutr 5:5140128. https://doi.org/10.1093/cdn/nzab035_036

Hannousse A (2021) Searching relevant papers for software engineering secondary studies: Semantic Scholar coverage and identification role. IET Softw 15:126–146. https://doi.org/10.1049/sfw2.12011

Indravadanbhai Patel C, Patel D, Patel Smt Chandaben Mohanbhai C, et al (2012) Optical Character Recognition by Open source OCR Tool Tesseract: A Case Study Digital ScareCrow using Iot View project CHARUSAT Apps (Mobile Application) View project Optical Character Recognition by Open Source OCR Tool Tesseract: A Case Study. Artic Int J Comput Appl 55:975–8887. https://doi.org/10.5120/8794-2784

Jiao W, Wang W, Huang J, et al (2023) Is ChatGPT a good translator? A preliminary study. arXiv Prepr arXiv230108745

Kanellopoulou A, Katelari A, Notara V, et al (2021) Parental health status in relation to the nutrition literacy level of their children: Results from an epidemiological study in 1728 Greek students. Med J Nutrition Metab 14:57–67. https://doi.org/10.3233/MNM-200470

Kasai J, Sakaguchi K, Le Bras R, et al (2022) Bidimensional Leaderboards: Generate and Evaluate Language Hand in Hand. NAACL 2022 - 2022 Conf North Am Chapter Assoc Comput Linguist Hum Lang Technol Proc Conf 3540–3557. https://doi.org/10.18653/v1/2022.naacl-main.259

Kocbek P, Gosak L, Musovic K, Stiglic G (2022) Generating Extremely Short Summaries from the Scientific Literature to Support Decisions in Primary Healthcare: A Human Evaluation Study. Artif. Intell. Med. AIME 2022 13263:373–382

Kojima T, Gu SS, Reid M, et al (2022) Large Language Models are Zero-Shot Reasoners

Korngiebel DM, Mooney SD (2021) Considering the possibilities and pitfalls of Generative Pre-trained Transformer 3 (GPT-3) in healthcare delivery. npj Digit Med 4:1–3. https://doi.org/10.1038/s41746-021-00464-x

Lee EK, Uppal K (2020) CERC: an interactive content extraction, recognition, and construction tool for clinical and biomedical text. BMC Med Inform Decis Mak 20:306. https://doi.org/10.1186/s12911-020-01330-8

Lin C-Y, Och FJ (2004) Looking for a few good metrics: ROUGE and its evaluation. In: Ntcir workshop

Liu Y, Iter D, Xu Y, et al (2023) GPTEval: NLG Evaluation using GPT-4 with Better Human Alignment. arXiv Prepr arXiv230316634

Luhn HP (1958) The Automatic Creation of Literature Abstracts. IBM J Res Dev 2:159–165. https://doi.org/10.1147/rd.22.0159

Mihalcea R (2004) Graph-based ranking algorithms for sentence extraction, applied to text summarization. 20-es. https://doi.org/10.3115/1219044.1219064

Moradi M, Blagec K, Haberl F, Samwald M (2021) GPT-3 Models are Poor Few-Shot Learners in the Biomedical Domain. https://doi.org/10.48550/arxiv.2109.02555

Nomoto T, Matsumoto Y (2001) A new approach to unsupervised text summarization. SIGIR Forum (ACM Spec Interes Gr Inf Retrieval) 26–34. https://doi.org/10.1145/383952.383956

OpenAI (2022) OpenAI. https://openai.com/. Accessed 1 Dec 2022

OpenAI (2023) Models - OpenAI API. https://platform.openai.com/docs/models/overview. Accessed 21 Jun 2023

Page MJ, McKenzie JE, Bossuyt PM, et al (2022) The PRISMA 2020 statement: an updated guideline for reporting systematic reviews. Rev Panam Salud Publica/Pan Am J Public Heal 46:105906. https://doi.org/10.26633/RPSP.2022.112

Radford A, Narasimhan K, Salimans T, Sutskever I (2018) Improving Language Understanding by Generative Pre-Training. Homol Homotopy Appl

Reunamo A, Peltonen LM, Mustonen R, et al (2022) Text Classification Model Explainability for Keyword Extraction-Towards Keyword-Based Summarization of Nursing Care Episodes. Stud Health Technol Inform 290:632–636. https://doi.org/10.3233/SHTI220154

Romera-Paredes B, Torr PHS (2015) An embarrassingly simple approach to zero-shot learning. 32nd Int Conf Mach Learn ICML 2015 3:2142–2151. https://doi.org/10.1007/978-3-319-50077-5_2

Sadeh-Sharvit S, Rego SA, Jefroykin S, et al (2022) A Comparison Between Clinical Guidelines and Real-

16	Next Generation eHealthWorld Treatment Data in Examining the Use of Session Summaries: Retrospective Study. JMIR Form Res 6:e39846. https://doi.org/10.2196/39846

Spaulding EM, Marvel FA, Piasecki RJ, et al (2021) User engagement with smartphone apps and cardiovascular disease risk factor outcomes: Systematic review. JMIR Cardio 5:. https://doi.org/10.2196/18834

Sreelekha S, Bhattacharyya P, Jha SK, Malathi D (2016) A survey report on evolution of machine translation. Int J Control Theory Appl 9:233–240

Stiglic G, Musovic K, Gosak L, et al (2022) Relevance of automated generated short summaries of scientific abstract: use case scenario in healthcare. 2022 IEEE 10TH Int. Conf. Healthc. INFORMATICS (ICHI 2022) 599–605

Stine JG, Soriano C, Schreibman I, et al (2021) Breaking Down Barriers to Physical Activity in Patients with Nonalcoholic Fatty Liver Disease. Dig Dis Sci 66:3604–3611. https://doi.org/10.1007/s10620-020-06673-w

Suanmali L, Salim N, Binwahlan MS (2009) Fuzzy Logic Based Method for Improving Text Summarization. IJCSIS) Int J Comput Sci Inf Secur 2:

Sutskever I, Vinyals O, Le Q V. (2014) Sequence to sequence learning with neural networks. Adv Neural Inf Process Syst 4:3104–3112

Trivedi G, Handzel R, Visweswaran S, et al (2018) An interactive NLP tool for signout note preparation. Proc. - 2018 IEEE Int. Conf. Healthc. Informatics, ICHI 2018 426–428

van der Lee C, Gatt A, van Miltenburg E, Krahmer E (2021) Human evaluation of automatically generated text: Current trends and best practice guidelines. Comput Speech Lang 67:101151. https://doi.org/10.1016/J.CSL.2020.101151

Vaswani A, Brain G, Shazeer N, et al (2017) Attention is All you Need. Adv Neural Inf Process Syst 30:

Wei J, Wang X, Schuurmans D, et al (2022) Chain-of-Thought Prompting Elicits Reasoning in Large Language Models

Yan R, Kong L, Huang C, et al (2011) Timeline generation through evolutionary trans-temporal summarization. In: EMNLP 2011 - Conference on Empirical Methods in Natural Language Processing, Proceedings of the Conference. pp 433–443

Zhang H, Cai J, Xu J, Wang J (2019) Pretraining-Based Natural Language Generation for Text Summarization. CoNLL 2019 - 23rd Conf Comput Nat Lang Learn Proc Conf 789–797. https://doi.org/10.48550/arxiv.1902.09243

Zhang J, Oh YJ, Lange P, et al (2020a) Artificial intelligence chatbot behavior change model for designing artificial intelligence chatbots to promote physical activity and a healthy diet: Viewpoint. J Med Internet Res 22:e22845. https://doi.org/10.2196/22845

Zhang J, Zhao Y, Saleh M, Liu PJ (2020b) PEGASUS: Pre-training with Extracted Gap-sentences for Abstractive Summarization